\documentclass[10pt,twocolumn,letterpaper]{article}

\usepackage[pagenumbers]{cvpr} 

\usepackage{graphicx}
\usepackage{amsmath}
\usepackage{amssymb}
\usepackage{booktabs}

\usepackage[pagebackref,breaklinks,colorlinks]{hyperref}

\usepackage[capitalize]{cleveref}
\crefname{section}{Sec.}{Secs.}
\Crefname{section}{Section}{Sections}
\Crefname{table}{Table}{Tables}
\crefname{table}{Tab.}{Tabs.}

\def\cvprPaperID{*****} 
\def\confName{CVPR}
\def\confYear{2022}

\begin{document}

\title{TopoMask: Instance-Mask-Based Formulation for the Road Topology Problem via Transformer-Based Architecture}

\author{M. Esat Kalfaoglu \and Halil Ibrahim Ozturk \and Ozsel Kilinc \and Alptekin Temizel \\
\and
Graduate School of Informatics, Middle East Technical University, Ankara, Turkey\\
{\tt\small esat.kalfaoglu@metu.edu.tr, atemizel@metu.edu.tr}
}

\maketitle

\begin{abstract}
\vspace{-0.2cm} 
    Driving scene understanding task involves detecting static elements such as lanes, traffic signs, and traffic lights, and their relationships with each other. To facilitate the development of comprehensive scene understanding solutions using multiple camera views, a new dataset called Road Genome (OpenLane-V2) has been released. This dataset allows for the exploration of complex road connections and situations where lane markings may be absent. Instead of using traditional lane markings, the lanes in this dataset are represented by centerlines, which offer a more suitable representation of lanes and their connections. 
    In this study, we have introduced a new approach called TopoMask for predicting centerlines in road topology. Unlike existing approaches in the literature that rely on keypoints or parametric methods, TopoMask utilizes an instance-mask based formulation with a transformer-based architecture and, in order to enrich the mask instances with flow information, a direction label representation is proposed. TopoMask have ranked $4^{th}$ in the OpenLane-V2 Score (OLS) and ranked $2^{nd}$ in the F1 score of centerline prediction in OpenLane Topology Challenge 2023. In comparison to the current state-of-the-art method, TopoNet, the proposed method has achieved similar performance in Frechet-based lane detection and outperformed TopoNet in Chamfer-based lane detection without utilizing its scene graph neural network. 
\end{abstract}
%
\vspace{-0.4cm}
\section{Introduction}
\vspace{-0.2cm}
\label{sec:intro}
Accurate detection of stationary objects plays a vital role in downstream tasks like planning and control in autonomous driving. Static elements on the road include ground objects, such as lanes, road lines, curbs, and crosswalks as well as above-ground objects like traffic lights, traffic signs, and poles. Currently, High-Definition Maps (HDMaps) are utilized for the safe autonomous driving. These maps provide pre-computed static-object maps. However, HDMaps are costly to extract and are typically available for limited areas. They cannot capture recent changes on the road, necessitating continuous maintenance. Additionally, Global Navigation Satellite System (GNSS) receiver on the vehicle can introduce localization errors. These errors, in turn, can cause discrepancies between the vehicle's actual position and the information provided by the HDMaps, potentially leading to inaccuracies and drifts in the map-based guidance system.

Automatic HDMap extraction has gained significant attention in recent years for two primary reasons, as highlighted by several studies \cite{li2022hdmapnet, liu2022vectormapnet, liao2022maptr, Can2021}. Firstly, it has a potential to eliminate the reliance on HDMaps for autonomous driving. Secondly, it can reduce the cost of extraction and continuous maintenance of HDMaps, minimizing the required human effort. The introduction of Road Genome dataset \cite{wang2023road}, which introduces the centerline concept to represent lanes, is an important milestone with regards to these aspects. Centerlines provide a more natural representation of lanes and bring a number of advantages over lane markings. Firstly, centerlines eliminate the need for post-processing to match lane dividers with their corresponding counterparts, simplifying the lane detection process. Secondly, they provide a more intuitive representation of the direction of traffic flow. Thirdly, centerlines capture the relationships between lanes more effectively, enhancing the understanding of lane connections. An additional benefit of centerlines is their ability to address situations where lane markings are absent, such as in road intersections.  In this dataset, association between lanes and traffic elements (taffic signs and lights) are also included into the problem definition which is significant for complete autonomous driving experience.  

In this study, we have introduced a novel approach called TopoMask for representing centerlines, departing from the key-point based and parametric methods such as bezier and polynomial approaches. TopoMask leverages a transformer-based architecture combined with an instance-mask-based representation. A similar segmentation-based approach is employed in HDMapNet \cite{li2022hdmapnet}. However, HDMapNet relies on semantic segmentation and requires additional instance embeddings and direction outputs, leading to a more complex post-processing step to merge the outputs. In contrast, our proposed formulation solely relies on the instance mask output with a novel direction label representation that denotes the flow of the centerlines: up, down, left and right. This simplifies the overall process and avoids the need for extensive post-processing.

To the best of our knowledge, TopoMask achieves the state-of-the-art performance in OpenLane-V2 dataset comparing with the existing methods in the literature. Remarkably, our approach achieves comparable Frechet-based centerline detection performance to TopoNet with a score of 22.8, even without incorporating a relation improver block such as TopoNet's Scene Graph Neural Network. In terms of Chamfer-based centerline detection performance, TopoMask significantly outperforms TopoNet with a score of 25.3.

In the OpenLane Topology Challenge 2023, our proposed architecture attains the $4^{th}$ position with a OLS score of 39.2 and its F1-score of 45.9 is the $2^{nd}$ highest score for centerline prediction.

\vspace{-0.3cm}
\section{Related Work}
\vspace{-0.2cm} 
Studies in the field of lane detection can be classified into four main branches: perspective methods, 3D lane divider based methods, multi-camera BEV methods, and centerline based methods. 

\vspace{-0.4cm}
\paragraph{Perspective Methods:}
These methods focus on detecting lane dividers from the perspective view, followed by projecting them onto the ground using a homography matrix under the assumption of a flat surface. Although road dividers are different instances, semantic approaches are well-suited for this task due to the constant number of lane divider instances (side-left, ego-left, ego-right and side-right). SCNN \cite{pan2018spatial} introduced a special module that applies sequential processes to the rows and columns of the feature map. The study in \cite{qin2020ultra} aims to improve the inference speed, and converts the formulation from pixel-based to grid-based and proposes row-based anchor formulation. LaneATT \cite{tabelini2021keep} developed an anchor concept specifically for lanes (road lines), drawing inspiration from object detection studies. The emergence of the CurveLanes dataset \cite{xu2020curvelane} led to the adoption of instance-segmentation-based methods. CondLaneNet \cite{liu2021condlanenet} introduced lane specific methodologies such as offset prediction and row-wise formulation. Some approaches such as Polynomial Order Structures \cite{tabelini2021polylanenet} and Bezier Curve Approaches \cite{feng2022rethinking} used polynomial or bezier curve representations to reduce post-processing efforts and improve curve learning.

\vspace{-0.4cm} 
\paragraph{3D Lane Divider methods:}
Recent advancements in lane detection have shifted focus towards directly predicting the 3D locations of lane instances, eliminating the need for post-processing on perspective outputs. This shift gained momentum with the introduction of 3D lane datasets such as such as Once-3DLanes \cite{yan2022once}, OpenLane \cite{chen2022persformer} and Apollo 3D Synthetic Lane dataset \cite{guo2020gen}. 

Persformer \cite{chen2022persformer} utilizes a deformable attention-based decoder that utilizes reference points obtained from Inverse Projection Mapping (IPM). It adapts the 2D anchor concept to 3D lane points. In the BEV-LaneDet method \cite{wang2022bev}, a grid-based approach is employed, where positive points belonging to the same instance are unified using an embedding concept with a triplet-based loss. It also incorporates a virtual camera concept to unify different extrinsics from multiple camera positions for the View Relation Module (VRM) \cite{pan2020cross}. PETRV2 \cite{Liu2022} extends its sparse query design for lane detection. M\^2-3DLaneNet explores the fusion of lidar and camera sensors and shows that these sensors are complementary in the 3D lane detection task. 

\vspace{-0.4cm} 
\paragraph{Multi-Camera BEV methods:}
HDMapNet \cite{li2022hdmapnet}, one of the first studies in this branch, aim to detect lane dividers, road dividers and pedestrian crossings by utilizing three different heads: semantic segmentation, instance embedding and direction head. However, the use of three heads necessitates substantial post processing to convert them into polyline or polygon output. VectorMapNet \cite{liu2022vectormapnet} adopts a two-stage coarse-to-fine approach. In the first stage, sparse anchor keypoints are detected. In the second stage, dense keypoints are detected by utilizing the outputs of the first stage in an autoregressive manner. Instead of following an autoregressive approach, MapTR \cite{liao2022maptr} predicts the points on the polyline or polygon directly with a permutation-invariant Hungarian matcher. InstaGraM \cite{shin2023instagram} models this problem as a graph and models vertices and edges as heatmap and distance transform, respectively in $8\times8$ patches. From each patch, an embedding is extracted by considering the vertex positions and local directional information obtained from the distance transform. Subsequently, a graph neural network structure is constructed using an attention-based method. The Sinkhorn algorithm is employed to iteratively normalize the exponential scores between the vertices.

\vspace{-0.4cm} 
\paragraph{Centerline Concept:}
STSU \cite{Can2021} is one of the earlier studies that predicts the centerlines instead of lane markings. An extention of this study introduces the concept of minimal cycles, which aims to reduce the area between two intersections \cite{can2022topology}. CenterLineDet \cite{xu2022centerlinedet} focuses on predicting centerlines from multiple cameras by leveraging temporal information. Additionally, it explores the impact of camera-LiDAR sensor fusion on centerline prediction. LaneGAP \cite{liao2023lane} considers centerlines as path-wise rather than piece-wise and instead of directly predicting centerline portions and their connections, the paths can be obtained by preprocessing the centerline graph. Subsequently, the model predicts the paths and converts them back into the piece-wise graph structure. TopoNet \cite{li2023topology} is a pioneering study that proposes an architecture which is specifically designed for the Road Genome project \cite{wang2023road}. It introduces a specific relation model that creates a graph neural network connecting centerlines and traffic elements. The predictions of centerlines are generated at the end of the topology network, which enhances centerline prediction accuracy. The incorporation of a dedicated traffic element embedding also improves the detection of centerline elements. 

\vspace{-0.3cm}
\section{Methodology}
\begin{figure}[t!]
  \centering
  \includegraphics[width=\linewidth]{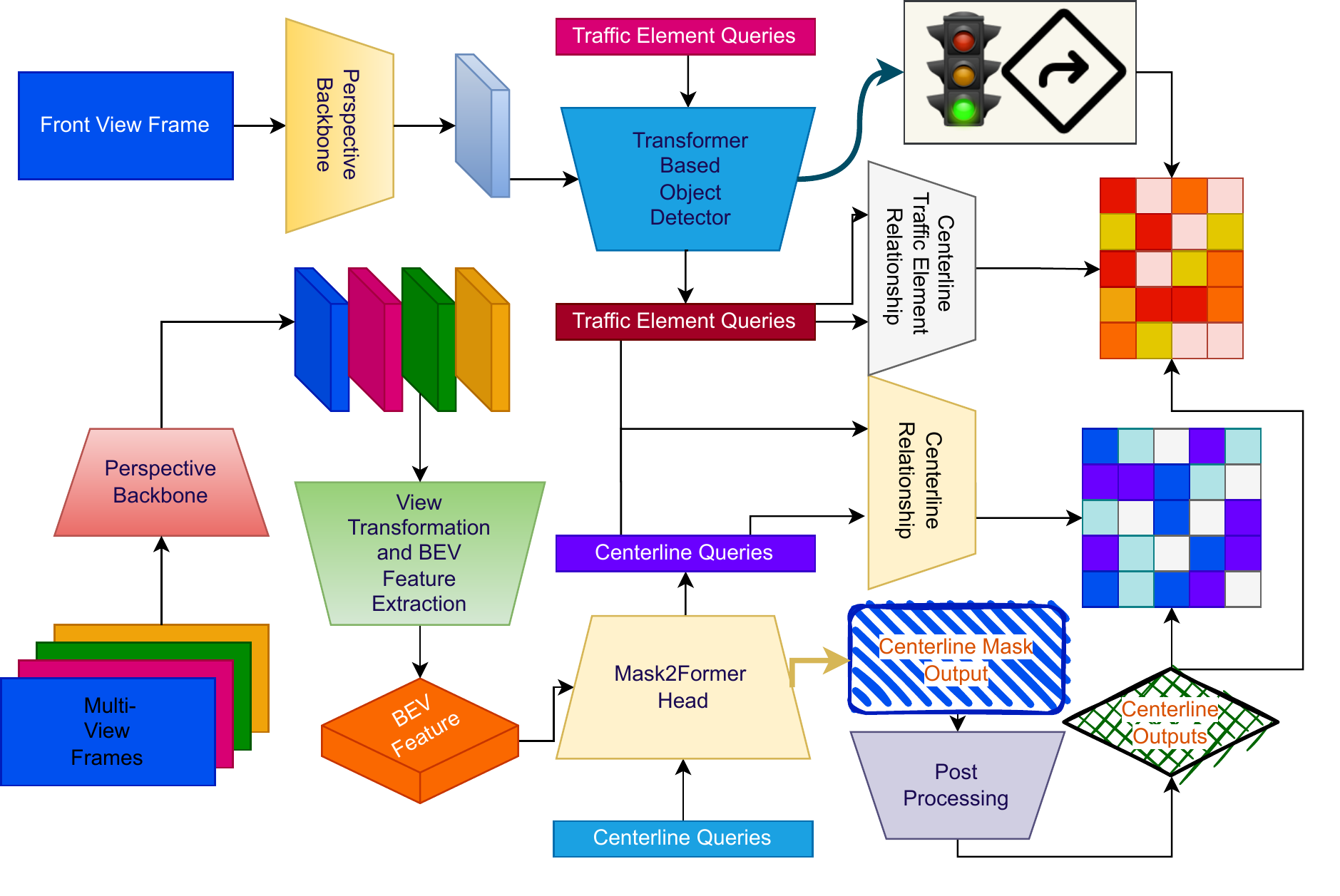}
  \vspace{-0.8cm} 
  \caption{Overview of the TopoMask architecture, featuring two main branches: one for detecting traffic elements and the other for detecting centerlines. The Mask2Former architecture treats the BEV feature as 2D image features, eliminating the need for additional structure. Queries from both transformer architectures are treated as embeddings and used in the relationship blocks. }
  \label{fig:topomask_architecture}
  \vspace{-0.4cm} 
\end{figure}

\vspace{-0.2cm} 
The general architecture of the TopoMask is shown in Figure \ref{fig:topomask_architecture}. In this formulation, both traffic element and centerline detection branches use transformer-based architectures. The centerline detection part of the TopoMask architecture relies on the instance-mask formulation. In this approach, each centerline instance is represented as a mask instance. By applying appropriate post-processing to each mask instance, it becomes possible to extract a point set for each proposal. 

\begin{figure}[t!]
  \centering
  \includegraphics[width=\linewidth]{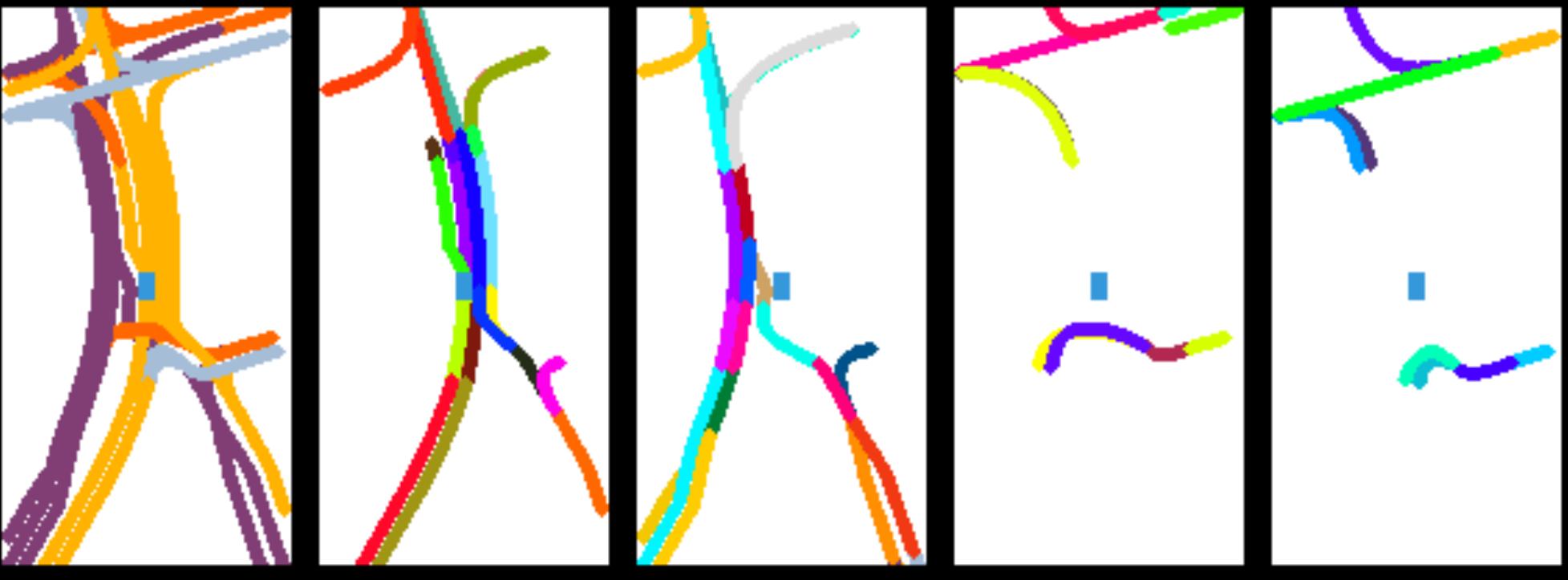}
  \vspace{-0.7cm} 
  \caption{Label representation of TopoMask: The leftmost image displays the semantic information of directions (up, down, left, and right) represented by distinct colors. The subsequent images depict the instances for each direction, showcasing the labels for up, down, left, and right, respectively. }
  \label{fig:label_representation}
  \vspace{-0.6cm} 
\end{figure}

However, the point set extracted from the mask instances does not have a specific order. In the problem definition of road topology, the flow information of each centerline is also crucial. In order to enrich the point sets with flow information, we propose a direction label representation (Figure \ref{fig:label_representation}) in TopoMask. As depicted in the leftmost image of the figure which represents the semantics of the direction information, there are four directions: up, down, left and right. The labels are assigned according to the \textit{dominant monotonicity} between the two axes of the ground truth point set. In the post-processing stage, according to the predicted direction label representation, point sets are ordered with respect to one axis ($x$ axis in vehicle coordinate system) for up and down direction labels, and ordered with respect to the other axis for left and right direction labels ($y$ axis in vehicle coordinate system). 

\vspace{-0.3cm}
\section{Experiments}
\vspace{-0.2cm}
\paragraph{Dataset and Metrics:} Argoverse 2 (AV2) \cite{argoverse2} dataset includes 7 cameras from 6 different cities and Subset-A release of the OpenLane-V2 dataset \cite{wang2023road} (also known as Road Genome) used in this study which is built on top of the AV2. Subset-A contains 22480 training, 4808 validation and 4816 test samples. For centerline and traffic element evaluation, the area considered is within +50 and -50 meters in the forward direction (x direction in the vehicle coordinate system) and +25 and -25 meters in the side direction (y direction in the vehicle coordinate system).

Both centerline detection and traffic element recognition utilize mAP (mean Average Precision) metrics for evaluation. True positive samples are determined using different distance measures depending on the task. For centerlines, the Frechet distance with thresholds of 1, 2, and 3 meters and Chamfer distance with 0.5, 1, and 1.5 meters are used. The Frechet distance takes into account both the distance and direction information between the predicted and ground truth centerlines while Chamfer distance ignores the direction information. For traffic elements, the Intersection over Union (IoU) metric is employed with a threshold of 0.75.

In addition to these metrics, a specific mAP metric has been proposed to evaluate topology reasoning in the graph domain. This metric takes into account the connectivity and relationships between centerlines and traffic elements. Note that, in order to assume connectivity (edge) as true positive, both vertices should be detected correctly regarding Frechet distance and IoU.  

F1-score is also calculated, but it is not included in the OLS metric. The F1-score is commonly used in the evaluation of predicted 3D lane dividers, such as in Gen-LaneNet \cite{guo2020gen}. According to this metric, if 75\% of the predicted points are within 1.5 meters error compared to the ground truth points of a corresponding instance, the sample is considered correct.

\vspace{-0.5cm}
\begin{equation}
\label{Eq:OLS}
    \text{OLS} = \frac{1}{4} \bigg[ \text{DET}_{l} + \text{DET}_{t} + f(\text{TOP}_{ll}) + f(\text{TOP}_{lt}) \bigg]
\vspace{-0.25cm}
\end{equation}

The evaluation system utilizes OpenLane-V2 Score (OLS) as the overall metric, which is calculated as the average of multiple task metrics (Eq. \ref{Eq:OLS}): centerline prediction with Frechet-based mAP ($\text{DET}_{l}$), traffic element prediction with IOU based mAP ($\text{DET}_{t}$), and topology relations between centerlines and traffic elements ($\text{TOP}_{ll}$ and $\text{TOP}_{lt}$). However, OLS does not take F1 score and Chamfer distance based mAP into account.

\vspace{-0.4cm}
\paragraph{Experimental Details:} The study utilizes DAB-DETR \cite{Liu2022_dab} for the 2D object detection and LSS \cite{philion2020lift} for the BEV feature map creation with $200\times104$ BEV resolution. On top of the BEV feature, Mask2Former architecture \cite{cheng2022masked} is utilized as instance mask prediction structure by preserving its original settings. For the relationship part, Sinkhorn iterations has been utilized as in InstaGraM study \cite{shin2023instagram}. To establish a directed graph for centerline relationships, the start and end points of the predictions are combined with the centerline embeddings in the form of sinusoidal positional encodings. The image resolution is $640\times896$ in BEV branch and a batch size of 16 is utilized. AdamW with a learning rate of 0.0003 is utilized as the optimizer with 0.1 multiplication on backbone gradients. As a post processing, expectation based location estimation is implemented as row-wise formulation for the up/down directions, and column-wise formulation for the left/right directions similar to \cite{qin2020ultra}. Then, second order polynomial fit is applied on top of the selected points and 11 points are sampled. During the mask prediction, bezier curves are also predicted with 5 control points in which start and end points are fixed. During bezier prediction, the 3D coordinates of centerlines are not normalized. 

\vspace{-0.4cm}
\paragraph{Experimental Results and Ablation Studies:}

We comparatively evaluate TopoMask with the current SOTA architecture, TopoNet, and report the results in Table \ref{tab:toponet-compare}. It should be noted that TopoMask does not have a relation improver method that enriches centerline embeddings with traffic element embeddings and implements second stage centerline prediction with these enriched embeddings. Despite this fact, TopoMask obtains on-par Frechet-based centerline detection performance and compares favorably in terms of Chamfer-based centerline detection and F1-scores. The results indicate that there is potential for improvement in the direction information of the TopoMask architecture. Additionally, the Frechet distance metric is negatively affected by sparse, highly erroneous detections among the 11-point predictions, particularly at intersections.  

\begin{table}[t!]
    \vspace{-0.2cm}
    \centering
    \scalebox{0.72}{
    	\begin{tabular}{l|cccccc}
    		\toprule
                Method & DET$_{l}$$\uparrow$ 
                & DET$_{l,\text{chamfer}}$$\uparrow$ 
                & TOP$_{ll}$$\uparrow$ & DET$_{t}$$\uparrow$ & TOP$_{lt}$$\uparrow$ 
                & OLS$\uparrow$ 
                \\
                \midrule
                TopoNet \cite{li2023topology} & 22.1 & 20.2 & 2.7 & 59.1 & 14.9 & 34.0\\
                TopoMask & 22.1 & \textbf{23.8} & \textbf{5.8} & 58.2 & \textbf{15.5} & \textbf{36.0}\\
                \bottomrule
    	\end{tabular}
    }
    \vspace{-0.2cm}
    \caption{Comparison of TopoMask and TopoNet architectures. Both architectures use ResNet-50 model as the backbone and both were trained for 24 epochs. The key distinction is that TopoMask does not utilize a shared feature extractor.}
    \label{tab:toponet-compare}
\end{table} 

\begin{table}[t!]
    \vspace{-0.2cm}
    \centering
    \scalebox{0.7}{
    	\begin{tabular}{l|ccc}
    		\toprule
                Config Type & BEV Backbone & 2D Backbone & \# Epochs 
                \\
                \midrule  
                Config1 & ResNet50 & ResNet50 & 25 \\
                Config2 & RegNetY-800mf & ResNet50 & 40 \\
                Config3 & RegNetY-800mf & ConvNextBase (OpenAi) & 40  \\ 
                Config3* & RegNetY-800mf & ConvNextBase (OpenAi) & 40 + 30 trainval  \\ 
                \bottomrule
    	\end{tabular}
    }
    \vspace{-0.2cm}
    \caption{Configs that are utilized in the ablations of Table \ref{tab:parametric_mask_comparsion}. }
    \vspace{-0.4cm}
    \label{tab:configs_ablations}
\end{table} 

\begin{table}[t!]
    \vspace{-0.2cm}
    \centering
    \scalebox{0.66}{
    	\begin{tabular}{l|cccccc}
    		\toprule
                Method & DET$_{l}$$\uparrow$ 
                & DET$_{l,\text{chamfer}}$$\uparrow$ 
                & TOP$_{ll}$$\uparrow$ & TOP$_{lt}$$\uparrow$  & OLS$\uparrow$ & F-score$\uparrow$
                \\
                \midrule
                Config1 Bezier & 14.4 & 14.7 & 4.2 & 11.4 & 31.7 & 36.1 \\
                Config1 Mask & \textbf{22.1} & \textbf{23.8} & \textbf{5.8} & \textbf{15.5} & \textbf{36.0} & \textbf{45.2} \\
                Config1 Fusion & 21.2 & 23.2 & 5.5 & 14.7 & 35.4 & 43.1 \\
                \hline
                Config2 Bezier & 17.7 & 20.1 & 4.9 & 12.1 & 33.1 & 40.1 \\
                Config2 Mask & \textbf{22.8} & \textbf{25.3} & 6.0 & 15.0 & 35.9 & \textbf{47.3} \\
                Config2 Fusion & 22.7 & 25.1 & 6.0 & \textbf{15.1} & \textbf{36.0} & 45.9 \\
                \hline
                Config3 Bezier & 19.4 & 20.9 & 5.2 & 13.2 & 35.0 & 41.3 \\
                Config3 Mask & \textbf{22.8} & \textbf{24.0} & 6.0 & 15.8 & 37.1 & \textbf{47.5} \\
                Config3 Fusion & 22.6 & 23.8 & \textbf{6.1} & \textbf{15.9} & 37.1 & 46.4 \\
                \hline
                Config3* Mask & 21.6 & NA & 5.8 & 15.6 & 38.9 & \textbf{46.1} \\
                Config3* Fusion & \textbf{22.1} & NA & \textbf{6.0} & \textbf{15.7} & \textbf{39.2} & 45.9 \\
                \bottomrule
    	\end{tabular}
    }
    \vspace{-0.2cm}
    \caption{Comparison of bezier parametric structure, mask structure and fusion strategies. Configurations vary in terms of backbones and training epochs and shown in Table \ref{tab:configs_ablations}. In some configurations, fusion options might increase the performance. * denotes the test results from the challenge. Test servers does not yield chamfer-based centerline detection performance for the test set, indicated as NA. }
    \vspace{-0.5cm}
    \label{tab:parametric_mask_comparsion}
\end{table}

TopoMask contains both mask prediction and parametric prediction in Bezier format. Therefore, potential fusion strategies are also explored. Table \ref{tab:parametric_mask_comparsion} shows the comparison of instance-mask and bezier based methodologies and their possible fusion. For the fusion, it is observed that accepting prediction for left/right direction from bezier predictions and up/down directions from mask predictions for some configurations is helpful. According to the results, masks have a superiority over bezier representation. 

\vspace{-0.3cm} 
\section{Conclusion and Future Work}
\vspace{-0.2cm}
In this work we introduced a novel instance-mask based methodology called TopoMask as an alternative to keypoint-based and parametric approaches commonly used in centerline detection. The performance of TopoMask was evaluated in comparison to the TopoNet architecture, and the results showed that TopoMask performs comparably to TopoNet in terms of Frechet-based centerline detection performance, while outperforming TopoNet in Chamfer-based centerline detection performance.

However, one limitation of the TopoMask architecture is its inability to predict the $z$ information. Assigning zero values to the ground truth $z$ leads to a Frechet-based detection score of approximately 0.92, indicating the potential for further improvement. Additionally, in the proposed architecture, some of the point sets of the centerlines are not monotonic, but the direction labels are assigned based on the dominant monotonicity of each axis. It is important to address these cases differently in future research. By applying the direction label strategy of TopoMask to the ground truths, a Frechet-based detection score of 0.92 is achieved, suggesting that there is still room for possible enhancement. These findings highlight the potential for refining the TopoMask methodology and addressing the mentioned limitations to further improve centerline detection performance. 

\vspace{-0.3cm}
\section*{ACKNOWLEDGMENTS}
\vspace{-0.2cm}
The numerical calculations reported in this paper were performed at TUBITAK ULAKBIM, High Performance and Grid Computing Center (TRUBA resources)

\vspace{-0.2cm}
{\small
\bibliographystyle{ieee_fullname}
\bibliography{references}
}

\end{document}